\documentclass[preprint,12pt,authoryear]{elsarticle}
\usepackage[utf8]{inputenc}
\usepackage[T1]{fontenc}

\usepackage{booktabs}
\usepackage{amsmath}    
\usepackage{amsthm}     
\usepackage{amssymb}    
\usepackage{stmaryrd}
\usepackage{natbib}     
\usepackage{subcaption}
\usepackage[hidelinks]{hyperref}
\usepackage[frozencache=true,cachedir=minted-cache,newfloat,outputdir=.]{minted}
\usepackage{txfonts}

\setminted[problog.py:ProbLogLexer -x]{
fontshape=n,
xleftmargin=20pt,
linenos=false,
escapeinside=@@,
mathescape=true,
breaklines=true,
}

\newcounter{lstlabelcounter}

\newminted[problog]{problog.py:ProbLogLexer -x}{}
\newcommand{\probloginline}[1]{\mintinline{problog.py:ProbLogLexer -x}{#1}}
\newcommand{\mathprobloginline}[1]{\text{\mintinline{problog.py:ProbLogLexer -x}{#1}}}

\SetupFloatingEnvironment{listing}{name=Listing}

\AtBeginEnvironment{minted}{}
\AtBeginEnvironment{problog}{}
\AtBeginEnvironment{problog*}{}

\DeclareMathOperator*{\argmin}{arg\,min}

\DeclareMathOperator*{\lpif}{\mathtt{{:}{\---}}}
\DeclareMathOperator*{\lpquery}{\mathtt{{?}{\---}}}
\DeclareMathOperator*{\prob}{\mathtt{{:}{:}}}

\newtheorem{definition}{Definition}
\newtheorem{example}{Example}

\newcommand{\theory}{\mathcal{T}}
\newcommand{\allmodels}[1]{\mathcal{M}(#1)}
\newcommand{\semiring}{\mathcal{S}}
\newcommand{\facts}{\mathcal{F}}
\newcommand{\rules}{\mathcal{C}}

\newcommand{\ive}[1]{\llbracket#1\rrbracket}

\newcommand{\differential}{\mathop{}\!{d}}

\journal{International Journal of Approximate Reasoning}

\begin{document}


\begin{frontmatter}



\title{Semirings for Probabilistic and Neuro-Symbolic\\Logic Programming \tnoteref{t1}}

\tnotetext[t1]{A peer-reviewed version of this preprint is published in the International Journal of Approximate Reasoning (\url{https://doi.org/10.1016/j.ijar.2024.109130})}


\author[kuleuven,leuvenAI]{Vincent Derkinderen\corref{cor1}}
\ead{vincent.derkinderen@kuleuven.be}
\cortext[cor1]{Corresponding author}
\author[kuleuven,leuvenAI]{Robin Manhaeve}
\ead{robin.manhaeve@kuleuven.be}
\author[sweden]{Pedro Zuidberg Dos Martires}
\ead{pedro.zuidberg-dos-martires@oru.se}
\author[kuleuven,leuvenAI,sweden]{Luc~De~Raedt}
\ead{luc.deraedt@kuleuven.be}

\affiliation[kuleuven]{organization={DTAI, Dept. of Computer Science, KU Leuven},
            city={Leuven},
            postcode={B-3000}, 
            country={Belgium}}

\affiliation[leuvenAI]{organization={Leuven.AI - KU Leuven Institute for AI},
            country={Belgium}}
            
\affiliation[sweden]{organization={Center for Applied Autonomous Systems, \"Orebro University},
            city={\"Orebro},
            postcode={SE-70182}, 
            country={Sweden}}

\begin{abstract}
   The field of probabilistic logic programming (PLP) focuses on integrating probabilistic models into programming languages based on logic.
   Over the past 30 years, numerous languages and frameworks have been developed for modeling, inference and learning in probabilistic logic programs.
   While originally PLP focused on discrete probability, more recent approaches have incorporated continuous distributions as well as neural networks,
   effectively yielding neural-symbolic methods. We provide a unified algebraic perspective on PLP, showing that many if not most of the 
   extensions of PLP can be cast within a common algebraic logic programming framework, in which facts are labeled with elements of a semiring and disjunction and conjunction are replaced by addition and multiplication. This does not only hold for the PLP variations itself but also for the underlying execution mechanism that is based on (algebraic) model counting.
\end{abstract}




\end{frontmatter}

\section{Introduction}\label{sec:intro}
 \cite{Poole:93} and \cite{Sato:95} 
 introduced the first frameworks for modeling, inference and learning probabilistic logic programming (PLP) in the early 90s. 
The original framework of Poole and Sato extended the logic programming language Prolog \citep{flach1994simplylogical} with probabilistic facts.
These are facts that are annotated with the probability that they are true; they play a role similar
to the parentless nodes in Bayesian networks in that they are marginally independent of one another,
and that the probabilistic dependencies are induced by the rules of the logic program.  This resulted in the celebrated
distribution semantics  \citep{Sato:95} that is the basis of probabilistic logic programming, and the corresponding
learning algorithm in the PRISM language  \citep{Sato:95} constitutes -- to the best of the authors' knowledge -- the very 
first probabilistic programming language with built-in support for machine learning. 
The work of Sato and Poole has inspired many follow-up works on inference and learning, and has also introduced many
 variations and extensions of the probabilistic logic programming and its celebrated distribution semantics.

Some of these works are concerned with faster inference using knowledge compilation technology (in ProbLog \citep{DeRaedt07}) 
and approximate inference \citep{Jonas} as well as 
extensions towards continuous distributions \citep{gutmann2011magic},  the use of semirings \citep{Eisner,KimmigBR11}, 
neural networks \citep{deepproblog,yang2020neurasp}, and dynamics \citep{Jonas}. An overview of many PLP techniques is provided in \citep{Riguzzi,DeRaedt15}.  

The key contribution of this paper is the insight that many of the extensions of the basic PLP framework
can be viewed within an algebraic logic programming framework \citep{Eisner,KimmigBR11}, in which the probabilistic semiring
is replaced by other, sometimes special purpose semirings. This holds, for instance, for working with discrete - continuous 
distributions in PLP as well as for computing gradients, for which the WMI semiring of \citep{zuidberg2019exact,zuidberg2023declarative} can be used
as well as the gradient semiring. This does not only hold, as we will show, at the level of logic programs, but also
at the circuit level when performing inference or learning. A circuit is obtained after grounding the logic program with regard to a specific query and compiling it into a boolean circuit \citep{Darwiche02}.  On the resulting circuit, one can then use semirings to compute probabilities,
integrals and gradients, which are useful for discrete and continuous inference as well as for learning. The resulting approach has been dubbed algebraic model counting \citep{kimmig2017algebraic}. A further insight
is that the algebraic approach allows to tightly couple neural networks to PLPs using the neural predicates as introduced in DeepProbLog \citep{manhaeve2021neural}.  These neural predicates are incorporated in PLP via so-called neural facts, which 
encapsulate a neural network computing a probability from its inputs.
Thus, this paper provides a general semiring framework for probabilistic and neural logic programming by 
allowing for other algebraic types of labels for the probabilistic facts than the probabilistic ones, which provides
a more general algebraic semantics and corresponding inference and parameter learning algorithm. In this way, it summarizes
the key insights obtained from our 15-year journey in PLP that started with Problog \citep{DeRaedt07}.
Throughout the paper we will focus on intuitions and on the simplest setting  for PLP based on labeled facts,
rather than exhaustively cover all (syntactic) variations for which 
 we will refer to the literature for technical details.

This paper is organised as follows:
in Section~\ref{sec:facts}, we give a brief introduction to logic programming, and how logic programs can be extended to a wide variety of domains such as statistical relational AI and neural-symbolic AI. We also show how these extensions are generalized by the concept of the algebraic fact and the use of semirings.
In Section~\ref{sec:inference}, we give an explanation of how inference happens for logic programming and its extensions.
In Section~\ref{sec:learning}, we show how learning happens for such programs.
Finally, we give an overview of related work and applications in Section~\ref{sec:related}, followed by concluding remarks in Section~\ref{sec:conclusion}.

\section{From Logic Programs to Algebraic Logic Programs}\label{sec:facts}
We will first introduce definite clause logic, which forms the basis of logic programming and the programming language Prolog.
Afterwards we will consider various variations in which facts are labeled, which are used in probabilistic and neural-symbolic logic programs.

\subsection{Logic programming}\label{sec:lp}
Logic programming is based on definite clauses. 
These are expressions of the form $h \lpif b_1, ... , b_N$ where $h$ and the $b_i$'s are logical atoms. A logical atom $a(t_1, ...,t_K)$ consists of a predicate $a$ of arity $K$ (often denoted $a/K$), followed by $K$ terms $t_i$. Terms then are either constants, logical variables or structured terms of the form $f(t_1, ... , t_L)$ with $f$ a functor and the $t_i$ terms. 
A clause of the form $h \lpif b_1, ... , b_N$ states that $h$ is true whenever all $b_i$'s are true.  When $N=0$, the clause is a fact and it is assumed to be true. 
A substitution $\theta$ is an expression of the form $\{V_1 = t_1, \dots , V_N = t_N\}$ where the $V_i$ are different variables and the $t_i$'s are terms. Applying a substitution $\theta$ to an expression $e$ (term or clause) yields the instantiated expression $e\theta$ where all variables $V_i$ in $e$ have been replaced by their corresponding terms $t_i$ in $e$. 
For example, applying the substitution $\theta=\{X=an, Y=bob\}$ to \probloginline{parent(X,Y)} results in \probloginline{parent(an,bob)}. 
When an expression does not contain any variables, it is called ground. 
Logic programs consist of two main components: 1) a set of facts $\facts$ that define the atoms that are considered true, and 2) a set of rules (or clauses) $\rules$ which allow the program to derive new atoms from the given set of facts through resolution. 
A logic program, together with a specific semantics defines the entailment relationship ( $\models$ ), which defines all the atoms that can be derived using the given facts and rules.
For further details on logic programming, we refer to \cite{flach1994simplylogical}.

Facts are a basic constituent of logic programs, they  represent atoms that are true. 
We will now show how they can be extended to cope with discrete, continuous, neural and algebraic labels
that form the basis of modern PLP.

\subsection{Probabilistic Facts}
\label{subsec:prob_facts}

The probabilistic fact is a generalisation of the logic fact, in which the fact is annotated with a probability of being true instead of being deterministically true. This lifts logic programming to probabilistic logic programming (PLP).

\begin{definition}[Probabilistic fact]
    A probabilistic fact is an expression of the form
$p :: f$
where $f$ is a ground fact that is true with a probability $p \in [0,1]$. 
\end{definition}

Introducing the probabilistic fact to the logic programming language Prolog resulted in the PLP language ProbLog \citep{DeRaedt07}.

\begin{definition}[ProbLog program]
    A ProbLog program is a triple $(\facts, w, \rules)$ where $\facts$ is a set of probabilistic facts, $w$ is a function mapping each ground probabilistic fact $f \in \facts$ to its probability $p$, and $\rules$ is a set of definite clauses.
\end{definition}

\begin{example}[Bayesian network]

\label{ex:bayesian_network}
The ProbLog program below models a variant of the well-known sprinkler Bayesian network. It contains three probabilistic facts and three rules.
\begin{problog*}{linenos}
0.25 :: cloudy.
0.8  :: humid.
0.5  :: sprinkler.

rain :- cloudy, humid.
wet  :- rain.
wet  :- sprinkler.
\end{problog*}
\end{example}
When considering all probabilistic facts in a program, the probability that a set of probabilistic facts $F' \subseteq \facts$ is true, and all other facts in the program ($\facts \setminus F'$) false, is given by~\citep{kimmig2017algebraic}:
\begin{equation}
    P_\facts(F') = \bigg(\prod_{f_i \in F'} w(f_i) \bigg) \bigg(\prod_{f_i \in \facts \setminus F'} (1-w(f_i)) \bigg)
\end{equation}

A set of facts $F' \subseteq \facts$ of a program, combined with the rules $\rules$ again form a deterministic program. In this way, the probabilistic facts induce a probability distribution over all the deterministic programs. The set of all atoms that are entailed to be true from $F'$ and $\rules$ together is often referred to as a \emph{possible world} \citep{kimmig2017algebraic}. 

\begin{example}[Possible worlds]
\label{ex:possible_worlds}
    Consider the ProbLog program in Example~\ref{ex:bayesian_network}. 
    The set of all ground probabilistic facts $\facts$ is 
    $\{\mathprobloginline{cloudy}, \mathprobloginline{humid}, \mathprobloginline{sprinkler} \}$. 
        An example $F'$ is $\{\mathprobloginline{cloudy}, \mathprobloginline{humid}\} \subset \facts$, which has probability
    \begin{equation}
        P_{\facts}(F') = (0.25 \times 0.8) \times (1-0.5) = 0.1
    \end{equation}
    The possible world entailed by $F'$ in the ProbLog program is the set
    \begin{equation}
        \{\mathprobloginline{cloudy},\; \mathprobloginline{humid},\; \mathprobloginline{rain},\; \mathprobloginline{wet}\}
    \end{equation}
    A different $F'$ would lead to a different possible world; the table below shows all the possible worlds of this ProbLog program along with their probability.
    
    {
    \centering
    \begin{tabular}{@{}lr@{}}
        \toprule
        Possible world & $P_{\facts}(F')$ \\ \midrule
        \{\} & 0.075 \\
        \{\probloginline{cloudy}\} & 0.025 \\
        \{\probloginline{humid}\} & 0.3 \\
        \{\probloginline{cloudy}, \probloginline{humid}, \probloginline{rain}, \probloginline{wet}\} & \textbf{0.1} \\
        \{\probloginline{sprinkler}, \probloginline{wet}\} & \textbf{0.075 }\\
        \{\probloginline{sprinkler}, \probloginline{cloudy}, \probloginline{wet}\} & \textbf{0.025} \\
        \{\probloginline{sprinkler}, \probloginline{humid}, \probloginline{wet}\} & \textbf{ 0.3} \\
        \{\probloginline{sprinkler}, \probloginline{cloudy}, \probloginline{humid}, \probloginline{rain}, \probloginline{wet}\} & \textbf{0.1} \\
        \bottomrule
    \end{tabular}\par
    }
\end{example}

\begin{definition}[Success probability]
\label{def:success_prob}
    The success probability $P(G)$ of a query $G$ considers all possible worlds in which $G$ is entailed, and sums their probabilities.
    \begin{equation}
        \label{eq:problog_wmc_semantics}
        P_{(\facts,w,\rules)}(G) = \sum_{\substack{F' \subseteq \facts \\ F' \cup \rules \; \models G}} P_{\facts}(F') 
        = \sum_{\substack{F' \subseteq \facts \\ F' \cup \rules \; \models G}} \bigg(\prod_{f_i \in F'} w(f_i) \bigg) \bigg(\prod_{f_i \in \facts \setminus F'} (1-w(f_i)) \bigg)
    \end{equation}
\end{definition}

\begin{example}
\label{ex:success_probability}
    The success probability for query $G=\mathprobloginline{wet}$ considers all possible worlds where $G$ is true (those with a bold probability in Example~\ref{ex:possible_worlds}'s table): $P(\mathprobloginline{wet}) = 0.1 + 0.075 + 0.025 + 0.3 + 0.1 = 0.6$
\end{example}

\subsection{Neural Facts}
Just like the probabilistic fact is a generalisation of the fact, the neural fact is a generalisation of the probabilistic fact. 
Instead of the probability being fixed, or treated as a single learnable parameter, the probability of the neural fact is parameterized by a neural network. 
This allows for the introduction of a neural probabilistic logic programming language, which is a type of neural-symbolic integration.
One of the simplest forms of the neural fact is where its probability is defined by a neural network binary classifier.
Introducing this neural fact to the ProbLog language leads to the introduction of DeepProbLog~\citep{manhaeve2021neural}.

\begin{definition}[Neural fact]
A neural fact is an expression of the form
\[
nn(m_r,[x_1,\dots,x_k]) \prob r(x_1,\dots,x_k).
\]
where $r$ is a predicate symbol, $nn$ is a reserved functor, $m_r$ uniquely identifies a neural network model that defines a probability distribution over the Boolean domain $\{true, false\}$, conditioned on the ground inputs to the neural network $x_1,\dots,x_k$. 
\end{definition}

The semantics of the neural fact is defined in terms of the semantics of regular probabilistic facts. A neural fact of the form $nn(m_r,[x_1,\dots,x_k]) \prob r(x_1,\dots,x_k)$ represents a probabilistic fact $f_{m_r}(x_1,\dots,x_k) \prob r(x_1,\dots,x_k)$ where $f_{m_r}$ is the function defined by network $m_r$.

\begin{example}[Neural fact]\label{ex:neuralfact}
Extending on Example~\ref{ex:bayesian_network}, we can use a neural network to predict whether the day will be cloudy, based on additional info that is provided, such as pressure and temperature. 
\begin{problog*}{linenos}
nn(cloudnet, [@$18^{\circ}C,998 hPa$@]) :: cloudy(@$18^{\circ}C,998 hPa$@).
0.8 :: humid.
0.5 :: sprinkler.

rain(T,P) :- cloudy(T,P), humid.
wet(T,P) :- rain(T,P).
wet(_,_) :- sprinkler.
\end{problog*}
We can now query this model for the probability of \probloginline{wet}, given a certain temperature and pressure: $P(\mathtt{\mathprobloginline{wet}(18^{\circ}C,998~hPa)})$.
\end{example}

The concept of the neural fact, which can be used to encode binary classifiers, can be extended to the neural annotated disjunction in order to encode multiclass classifiers. For more details, we refer to \cite{manhaeve2021neural}.

\subsection{Distributional Facts and Indicator Facts}

As ProbLog and DeepProblog only allow for (neural) probabilistic facts, they are inherently restricted to discrete random variables.
We alleviate this by introducing so-called distributional facts and indicator facts\footnote{\citet{zuidberg2023declarative} implicitly introduced indicator facts by means of Boolean comparison predicates.} \citep{zuidberg2023declarative}.

\begin{definition}[Distributional Fact]
A distributional fact is of the form $x \sim d$, with $x$ being a ground term, and $d$ a ground term whose functor denotes a probability distribution.
The distributional fact states that the ground term $x$ is a random variable distributed according to $d$.
\end{definition}

\begin{definition}[Indicator Fact]
An indicator fact is an expression of the form
$\sigma :: f$
where $f$ is a ground fact labeled with a measurable set $\sigma$. 
\end{definition}

\begin{example}
\label{ex:bayesian_network_distributional}
 We rewrite the program  in Example \ref{ex:bayesian_network} using distributional and indicator facts. Note how the program separates into two layers. On the one hand we have the distributional facts that tell us how random variables are distributed. On the other hand we have logical rules.
 The link between the two layers is made by the indicator facts that 
 are each labeled with a measurable set using the random variables introduced by the distributional facts.
\begin{problog*}{linenos}
@$x_c$@ ˜ flip(0.25).
@$x_h$@ ˜ flip(0.8).
@$x_s$@ ˜ flip(0.5).

[@$x_c=1$@]::cloud.
[@$x_h=1$@]::humid.
[@$x_s=1$@]::sprinkler.

rain :- cloudy, humid.
wet  :- rain.
wet  :- sprinkler.
\end{problog*}
\end{example}

\begin{example}
\label{ex:bayesian_network_distributional_beta}

In Example \ref{ex:bayesian_network} and Example \ref{ex:bayesian_network_distributional} we modeled the humidity as a Boolean random variable. 
Extending ProbLog with continuous random variables allows us now to model the humidity as a continuous variable. Using, for instance, a beta distribution, we model the relative humidity as a random variable that takes values in the $[0,1]$ interval. 

\begin{problog*}{linenos}
@$x_c$@ ˜ flip(0.25).
@$x_h$@ ˜ beta(4,2).
@$x_s$@ ˜ flip(0.5).

[@$x_c=1$@]::cloud.
[@$x_h>0.6$@]::humid. @\label{line:indicator:humidity}@
[@$x_s=1$@]::sprinkler.

rain :- cloudy, humid.
wet  :- rain.
wet  :- sprinkler.
\end{problog*}
As we model the humidity as a continuous random variables we use an inequality instead of an equality in Line \ref{line:indicator:humidity}.
\end{example}

In order to define the success probability of a probabilistic program with distributional and indicator facts we resort to measure theoretic concepts.
While this introduces some notational overhead, defining the success probability for arbitrary probability distributions (not only Bernoulli distributions) becomes straightforward. In other words, this allows us to extend probabilistic logic programming from the discrete domain to the discrete-continuous domain. 
\begin{definition}[Measure theoretic success probability]
\label{def:measure_success_prob}
    The success probability $P(G)$ of a query $G$ is given by
    \begin{equation}
        \label{eq:problog_wmc_semantics_measure}
        P_{(\facts,\sigma,\rules)}(G) = 
        \int \left(
            \sum_{\substack{F' \subseteq \facts \\ F' \cup \rules \; \models G}}
            \left(
                \prod_{f_i \in F'} \sigma(f_i) 
            \right)
            \left(
                \prod_{f_i \in \facts \setminus F'} \sigma(\neg f_i)
            \right)
        \right)
        \differential \mu
    \end{equation}
    where $\sigma(\cdot)$ is a function that maps indicator facts to measurable indicator functions and $\differential \mu$ denotes the differential of the probability measure induced by the set of distributional facts.
\end{definition}

\begin{example}[Indicator functions]
\label{ex:sigma}
The function $\sigma(\cdot)$ in Definition \ref{def:measure_success_prob} maps the the indicator facts from Example \ref{ex:bayesian_network_distributional} to indicator functions as follows:
\begin{align*}
\begin{split}
&\sigma(\mathprobloginline{cloud}) = \ive{x_c=1} 
\\
&\sigma(\mathprobloginline{humid}) = \ive{x_h>0.6} 
\\
&\sigma(\mathprobloginline{sprinkler}) = \ive{x_s=1} 
\end{split}
\quad
\begin{split}
&\sigma(\neg\mathprobloginline{cloud}) = \ive{x_c=0}
\\ 
&\sigma(\neg\mathprobloginline{humid}) = \ive{x_h\leq 0.6} 
\\
&\sigma(\neg\mathprobloginline{sprinkler}) = \ive{x_s=0} 
\end{split}
\end{align*}
Here we use Iversion brackets $\ive{\cdot}$ do denote the indicator function: whenever the relation inside an Iverson brackets holds it evaluates to 1. It evaluates to 0 otherwise.
\end{example}

\begin{example}[Differential of probability measure]
\label{ex:dmu}
The distributional facts in Example \ref{ex:bayesian_network_distributional_beta} induce a probability measure that we can write as a product of measures, and we can also write the differential as a product of differentials:
$$\differential \mu =  \differential \mu_c \times \differential \mu_h \times \differential \mu_s$$
Using the functional form of the Bernoulli and beta distribution we can rewrite the measure explicitly in terms of the random variables $x_c$, $x_h$, and $x_s$:
\begin{align*}
\differential \mu
=
\Bigg(
0.25^{x_c} (1-0.75)^{x_c}
\differential x_c
\Bigg)
\Bigg(
\frac{1}{B(4,2)} x_h^3 (1-x_h)^1 
\differential x_h
\Bigg)
\Bigg(
0.5^{x_s} (1-0.5)^{x_s}
\differential x_s
\Bigg)
\end{align*}
The symbol $B(\cdot, \cdot)$ denotes the beta function, used to normalize the beta distribution.
\end{example}

\begin{example}
    
Using Definition \ref{def:measure_success_prob} we can compute the success probability for the query $G=\mathprobloginline{wet}$ in a similar fashion to Example \ref{ex:success_probability}. For the world
$
\{
\mathprobloginline{cloudy},\allowbreak
\mathprobloginline{humid},\allowbreak
\mathprobloginline{rain},\allowbreak
\mathprobloginline{wet}  
\}
$
we get:

\begin{align*}
    \int \ive{x_c=1} \ive{x_h>0.6} \ive{x_h=0} \differential \mu
\end{align*}
Plugging now in the expression for the differential in Example \ref{ex:dmu} and rearranging the factors the integrals we get:
\begin{align*}
    &\left(
    \int
    \ive{x_c=1}
    0.25^{x_c} (1-0.75)^{x_c}
    \differential x_c
    \right)
    \times 
    \left(
    \int
    \ive{x_h>0.6}
    \frac{1}{B(4,2)} x_h^3 (1-x_h)^1 
    \differential x_h
    \right)
    \\
    & \quad \times
    \left(
    \int
    \ive{x_s=0}
    0.5^{x_s} (1-0.5)^{x_s}
    \differential x_s
    \right)
    \\
    &=
    0.25 \times 0.66304 \times 0.5 = 0.08288
\end{align*}

 We can compute in a similar fashion the probabilities of the remaining worlds.
\end{example}

The program in Example \ref{ex:bayesian_network_distributional_beta} is rather simple in terms of functional dependencies between random variables. For instance, none of the parameters of any of the distributions depends on the parameters of another random variable. Furthermore, the labels of the indicator facts are all univariate. This means, effectively, that all the random variables are independent of each other. Note however that Definition \ref{def:measure_success_prob} is more general and does allow for functional dependencies between random variables and also for multivariate labels of indicator facts. We refer the interested reader to \cite{zuidberg2023declarative} for an in depth and formal exposition of such cases. We also note that \citet{de2023neural} generalize distributional facts to so-called {\it neural distributional facts}, which allows them to unify neural and discrete-continuous PLP. The main idea is that parameters of distributional facts are allowed to be the output of neural networks.

\subsection{Algebraic Facts}

The previously introduced concepts of probabilistic fact, neural fact, and indicator fact can each be generalised into an \emph{algebraic fact}, a concept first discussed in aProbLog~\citep{KimmigBR11}.
\begin{definition}[Algebraic fact]
    An algebraic fact is an expression of the form $a :: f$ where $f$ is a fact and $a$ is an element of a commutative semiring's domain.
\end{definition}

\begin{definition}[Commutative semiring] \citep{derkinderen2020algebraic}
    A commutative semiring $\mathcal{S}$ is an algebraic structure $(\mathcal{A}, \oplus, \otimes, e^\oplus, e^\otimes)$ where 
    \begin{itemize}
        \setlength\itemsep{0em}
        \item $\mathcal{A}$ defines the domain of the values,
        \item $\oplus$ and $\otimes$ are associative, commutative binary operations over $\mathcal{A}$,
        \item $\otimes$ distributes over $\oplus$,
        \item $e^\otimes \in \mathcal{A}$ and $\forall a \in \mathcal{A}\colon e^\otimes \otimes a = a$, i.e. $e^\otimes$ is a neutral element for $\otimes$.
        \item $e^\oplus \in \mathcal{A}$ and $\forall a \in A\colon e^\oplus \oplus a = a$ and $e^\oplus \otimes a = e^\oplus$, i.e. $e^\oplus$ is a neutral and absorbing element for $\oplus$ and $\otimes$ respectively.
    \end{itemize}
\end{definition}

\begin{example}[Commutative semiring]
    Examples of relevant commutative semirings include the Boolean semiring $(\mathbb{B},\lor,\land,\bot, \top)$, the probability semiring $(\mathbb{R}, +,\allowbreak \times, 0, 1)$, and the most probable explanation semiring $(\mathbb{R}, max, \times, 0, 1)$.
\end{example}

In this generalisation a program has four components $(\facts,\alpha,\rules,\semiring)$ where $\semiring$ is a commutative semiring and $w$ became $\alpha$ mapping not only facts $f$ but also their negation $\neg f$ to elements of $\semiring$'s domain $\mathcal{A}$. This allows $\neg f_i$ to be associated to something else than $1 - w(f_i)$. The inference task within such a program is generalised to Equation~\ref{eq:algebraic_fact_semantics}. Note the use of the semiring operations $\otimes$ and $\oplus$, which replaced $\times$ and $+$. We also replaced $P_{(\facts,\alpha,\rules)}(G)$ with $AMC_{(\facts,\alpha,\rules,\semiring)}(G)$ since it is no longer necessarily a probability.
\begin{equation}
    \label{eq:algebraic_fact_semantics}
    AMC_{(\facts,\alpha,\rules,\semiring)}(G) = 
    \bigoplus_{\substack{F' \subseteq \facts \\ F' \cup \rules \; \models G}} \bigg(\bigotimes_{f_i \in F'} \alpha(f_i) \bigg) \bigg(\bigotimes_{f_i \in \facts \setminus F'} \alpha(\neg f_i) \bigg)
\end{equation}

\begin{example}
    Consider the program in Example~\ref{ex:bayesian_network}. When using $(\mathbb{R},max,\times,0,1)$ as a semiring the output of the query becomes instead the most probable explanation. Indeed, $\otimes = \times$ means that the weight of a model $m$ is still its probability, and by choosing $\oplus = max$ the most probable model is selected.
\end{example}

While the example above is also solving a probabilistic task, the algebraic framework is certainly not restricted to probabilities alone. In fact, the semiring elements $\mathcal{A}$ can be anything as long as we define the proper operations over them: preference values, distances, weights, tuples, sets, \dots

Several other extensions built around this framework include reasoning over second-order queries~\citep{VerreetDMR22}, decision making via the expected utility semiring~\citep{derkinderen2020algebraic}, and parameter learning via the gradient semiring~\citep{KimmigBR11}. More information on the latter is provided in Section~\ref{sec:learning}.
Comparing Equation \ref{eq:algebraic_fact_semantics} to Equation \ref{eq:problog_wmc_semantics_measure} we also see that we can formulate the probability of a query to a discrete-continuous probabilistic program in terms of an algebraic model count as well:
\begin{align*}
    P_{(\facts,\sigma,\rules)}(G) =
    \int AMC_{(\facts,\sigma,\rules,\semiring)}(G) \differential \mu
\end{align*}
A similar formulation is also used in \citep{zuidberg2019exact,miosic2021measure,zuidberg2023declarative,de2023neural}.

\section{Inference}\label{sec:inference}

We now discuss inference for algebraic logic programs. Inference happens in three steps: 1) logical inference, 2) translation to an algebraic model counting problem, 3) calculating the algebraic model count. We now discuss each step in detail.

%
%
\subsection{Logical inference} \label{sec:logical_inference}
The first step concerns the logical inference, for which we discern two different but related approaches. 

The first approach is proving, which uses SLD resolution to calculate the set of all proofs for a query $G$.
SLD resolution uses a backward chaining, goal-oriented approach. 
A goal is a sequence of atoms $\lpquery l_1, ..., l_n$. 
The initial goal is the query. At each step, the algorithm chooses a clause $h \lpif ~b_1, \dots ,b_n$ whose head $h$ unifies with the first atom $l_1$ in the goal, with the substitution $\theta$, i.e. $h\theta = l_1\theta$. 
The application of resolution yields a new goal, $\lpquery (b_1, \dots b_n, l_2, \dots,  l_j)\theta$. 
This is repeated until the goal is empty, resulting in a successful proof, or until no more clauses can be applied, in which case the proof fails. 
It is possible that multiple clauses can be applied to a goal, which leads to different branches in the SLD tree and the possibility of multiple proofs for a single query.

The second approach is to construct the relevant ground program $\mathcal{P}_G$. 
Grounding replaces each rule $c$ containing variables $\{V_1, ... , V_k\}$ by all instances  $ c\theta$ where $\theta$ is a substitution $\{V_1 = c_1, ... V_k = c_k\}$ and the $c_i$ are constants or other ground terms appearing in the domain. 
If $G$ is not ground, the grounding will compute all possible answer substitutions $G\theta$.
To keep inference tractable, it is key to only consider that part of the ground program that is relevant to the query (i.e. the grounded facts and rules are used in the derivations of the query).
Again, SLD resolution is used to find the relevant grounding.

\begin{example}[Logical inference] \label{ex:inference}
We demonstrate the different steps of inference by extending Example~\ref{ex:bayesian_network} to reason about separate days of the week. 
For ease of modeling (and grounding), we also use rules annotated with probabilities, a purely syntactical construct \footnote{For example, the rule \probloginline{0.25 :: cloudy(sunday) :- day(sunday)} is syntactic sugar for a rule \probloginline{cloudy(sunday) :- cloudy_on_day(sunday), day(sunday)} and a fact \probloginline{0.25 :: cloudy_on_day(sunday)}.}.

\begin{problog*}{linenos}
day(monday).
...
day(sunday).

0.25 :: cloudy(Day) :- day(Day).
0.5 :: sprinkler(Day) :- day(Day).

0.8 :: rain(Day) :- cloudy(Day).

wet(Day) :- rain(Day).
wet(Day) :- sprinkler(Day).
\end{problog*}
We query the probability of the grass being wet on Sunday, i.e. $P(\mathprobloginline{wet(sunday)})$. We first consider the proving approach. The proving procedure is easily visualized as an SLD tree, shown in Figure~\ref{fig:sld}. There are two proofs for our query, one where the grass is made wet by the sprinkler, and one where it was rainy and cloudy.

\begin{figure}
    \centering
    \includegraphics[width=0.5\linewidth]{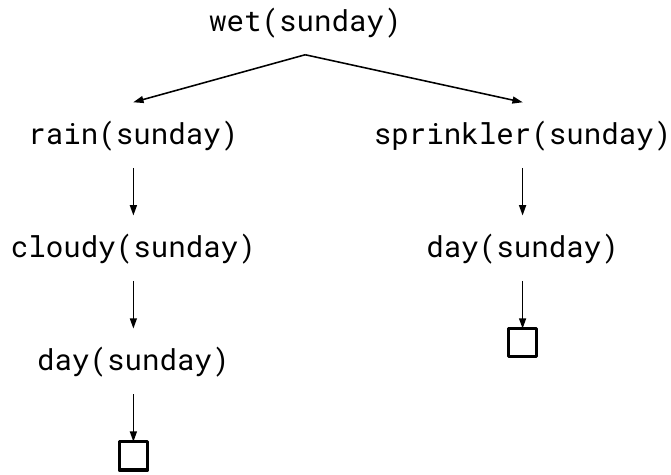}
    \caption{The SLD tree for Example~\ref{ex:inference}. The two branches represent the two separate proofs for the query \probloginline{wet(sunday)}.}
    \label{fig:sld}
\end{figure}

For the grounding approach, we only consider the part that is relevant to the query $\mathtt{wet(sunday)}$, so the variable $\mathtt{Day}$ only needs to be substituted with $\mathtt{sunday}$. The resulting relevant ground program is:
\begin{problog*}{linenos}
day(sunday).

0.25 :: cloudy(sunday) :- day(sunday).
0.5 :: sprinkler(sunday) :- day(sunday).

0.8 :: rain(sunday) :- cloudy(sunday).

wet(sunday) :- rain(sunday).
wet(sunday) :- sprinkler(sunday).
\end{problog*}
\end{example}

%
%
\subsection{Translation to algebraic model counting}
In the next step of the inference we map our results onto propositional logical formulas, for which we use the following concepts and notations.
A literal $l$ is an atom $v$ or its negation $\neg v$. A truth assignment to each atom is called an interpretation, which we represent as a set of literals $M$ where $v \in M$ iff $v$ is assigned to true and $\neg v \in M$ otherwise. An interpretation is called a model $M$ of a theory $\theory$ when $\theory$ is satisfied in $M$, formally denoted as $M \models \theory$. We use $\allmodels{\theory}$ to denote the set of all models of a theory $\theory$. 

%
%
Note that the possible worlds of a ProbLog program discussed in Section~\ref{subsec:prob_facts} are simply models of a theory formed by that program. Considering this equivalence, the probability query $P(G)$ (Equation~\ref{eq:problog_wmc_semantics}) can be identified as an instance of the weighted model counting problem. Consequently the problem of computing $P(G)$ can be addressed by transforming it into a weighted model counting problem for which several solvers exist.

\begin{definition}[weighted model count] The weighted model count (WMC) of a propositional logic theory $\theory$ over variables $\mathbf{V}$ and a weight function $\alpha$ mapping literals of $\mathbf{V}$ to a real, is 
\begin{equation}
    WMC(\theory, \alpha) = \sum_{m  \in \allmodels{\theory}} \prod_{l \in m} \alpha(l) \label{eq:def_WMC}
\end{equation}
 where 
 a model $m$ is a set of its literals such that $\prod_{l \in m} \alpha(l)$ is the total weight of $m$~\citep{chavira2008probabilistic}. We use $WMC(\theory)$ to refer to the WMC polynomial of $\theory$ where $\alpha$ is unspecified.
\end{definition}

\begin{example}[weighted model count]\label{ex:wmc}
Consider the propositional theory $\theory$ and weight function $\alpha$ below.
The models of $\theory$, denoted as $\mathcal{M}(\theory)$, correspond exactly to the possible worlds of Example~\ref{ex:bayesian_network}. Additionally, because we have chosen $\alpha$ appropriately, the weighted model count is the probability of \probloginline{wet} being true, $WMC(\theory,\alpha) = 0.6$.
\begin{align*}
    \theory = 
        \big( &\mathprobloginline{rain} \iff \mathprobloginline{cloudy} \land \mathprobloginline{humid} \big) \land 
        \big( \mathprobloginline{wet} \iff \mathprobloginline{rain} \lor \mathprobloginline{sprinkler} \big) \\
    \alpha = \{
        &\mathprobloginline{cloudy} \mapsto 0.25, 
        \neg\mathprobloginline{cloudy} \mapsto 0.75,
        \mathprobloginline{humid} \mapsto 0.8, 
        \neg\mathprobloginline{humid} \mapsto 0.2, \\
        &\mathprobloginline{sprinkler} \mapsto 0.5, 
        \neg\mathprobloginline{sprinkler} \mapsto 0.5,
        \mathprobloginline{rain} \mapsto 1, 
        \neg\mathprobloginline{rain} \mapsto 1, \\
        &\mathprobloginline{wet} \mapsto 1, 
        \neg\mathprobloginline{wet} \mapsto 0
    \}
\end{align*}
\end{example}

%
%
Similarly, the algebraic query (Equation~\ref{eq:algebraic_fact_semantics}) can be identified as an instance of algebraic model counting.

\begin{definition}[algebraic model count] 
     The algebraic model count (AMC) of a propositional logic theory $\theory$ over variables $\mathbf{V}$, a commutative semiring $\semiring = (\mathcal{A}, \oplus, \otimes, e^\oplus, e^\otimes)$, and a labeling function $\alpha$ mapping literals of $V$ to $a \in \mathcal{A}$, is \citep{kimmig2017algebraic}
    \begin{equation}
        AMC(\theory,\semiring,\alpha) = \bigoplus_{m \in \allmodels{\theory}}\bigotimes_{l \in m} \alpha(l) \label{eq:def_AMC}
    \end{equation}
\end{definition}

\begin{example}[algebraic model count] 
    \citep{derkinderen2020algebraic} Consider Example~\ref{ex:wmc} but using semiring $\semiring=(\mathbb{R},max,\times,0,1)$, then $AMC(\theory,\semiring,\alpha) = 0.3$, the highest model weight out of all $m \in \allmodels{\theory}$.
\end{example}

The construction of the logical formula itself depends on how the previous inference step was performed.
In the proving approach, the proofs for a query $q$ are combined into a logical formula
\[
G \leftrightarrow \bigvee_{E \in \text{Proofs}(G)} \bigwedge_{f_i \in E} f_i
\]
where the $f_i \in E$ are the probabilistic facts used in proof $E$.

In the grounding approach, Clark's completion can be used for cycle-free programs.
Clark's completion constructs a formula
\[
h \leftrightarrow \bigvee_{(h \lpif b_1,...,b_n) \in \mathcal{P}_G}b_1 \wedge \dots \wedge b_n 
\]
for each set of rules with the same ground head $h$, whose bodies are $b_1 , \dots, b_n$. 
Cyclical programs first need to be turned into equivalent acyclic programs through cycle breaking. For this, we refer to \citet{fierens2015inference}. The propositional theory $\theory$ in Example~\ref{ex:wmc} would for instance follow from this grounding approach.

%
%
\subsection{Solving model counting}

Weighted model counting is a \#P-complete problem~\citep{Valiant92}. Fortunately, distributivity, associativity and commutativity can often be exploited to drastically reduce the number of required computations. The state of the art approach that we highlight recognizes that theory $\theory$ dictates the computation and instead takes a more logical perspective. More specifically, it first compiles $\theory$ into a representation that then facilitates an efficient counting operation in a second phase. 
Compilation algorithms are studied in the knowledge compilation research domain, and they aim to produce succinct suitable representations fast~\citep{darwiche02knowledge, darwiche2011sdd}.

A suitable target representation language from which counting is efficient is \textit{sd-DNNF}. 
Adapted from~\cite{kimmig2017algebraic},
\begin{definition}[sd-DNNF]
    A propositional theory $\theory$ is in negation normal form (NNF) when the only Boolean operations in $\theory$ are $\{\neg, \land, \lor\}$, and when negation ($\neg$) only occurs on variables.
    $\theory$ is in sd-DNNF when it is in NNF and has the following additional properties:
    \begin{itemize}
        \item smooth (s): when for each disjunction node $\bigvee_{i=1}^n \phi_i$, each child $\phi_i$ mentions the same set of variables.
        \item deterministic (d): when for each disjunction $\bigvee_{i=1}^n \phi_i$, each pair of different children $\phi_i$ and $\phi_j$ is logically inconsistent.
        \item decomposable (D): when for each conjunctions $\bigwedge_{i=1}^n \phi_i$, no two children $\phi_i$ and $\phi_j$ share any variables.
    \end{itemize}
\end{definition}

%
%
This representation\footnote{A d-DNNF is sufficient, as smoothing can be performed during evaluation.} can easily be converted into a computational graph that represents the weighted model count polynomial $WMC(\theory)$. 
Simply replace each $\lor$ into $+$, each $\land$ into $\times$, each literal $l$ into $\alpha(l)$, each true ($\top$) into $1$, and each false ($\bot$) into $0$.
The soundness of this procedure can be proven inductively, we refer to \cite{kimmig2017algebraic}. To provide a high-level intuition: determinism ensures that each model is considered at most once, decomposability ensures each literal is considered at most once, and smoothness is important when $\alpha(l) + \alpha(\neg l) \neq 1$. The computational graph resulting from this procedure is called an \emph{arithmetic circuit}~\citep{Darwiche02}.

\begin{figure}
    \centering
    \includegraphics[width=\linewidth]{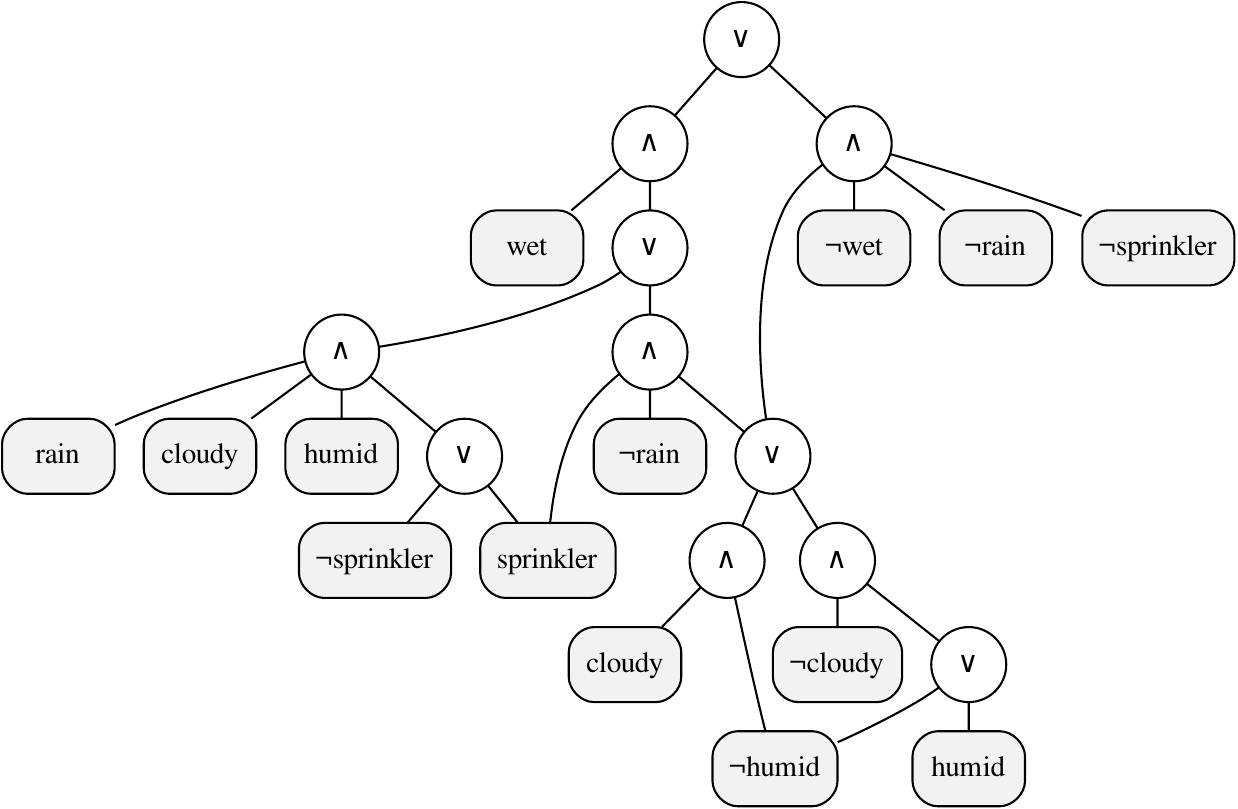}
    \caption{An sd-DNNF corresponding to the ProbLog program in Example~\ref{ex:bayesian_network}.}
    \label{fig:sd_ddnnf}
\end{figure}

\begin{figure}
    \centering
    \includegraphics[width=\linewidth]{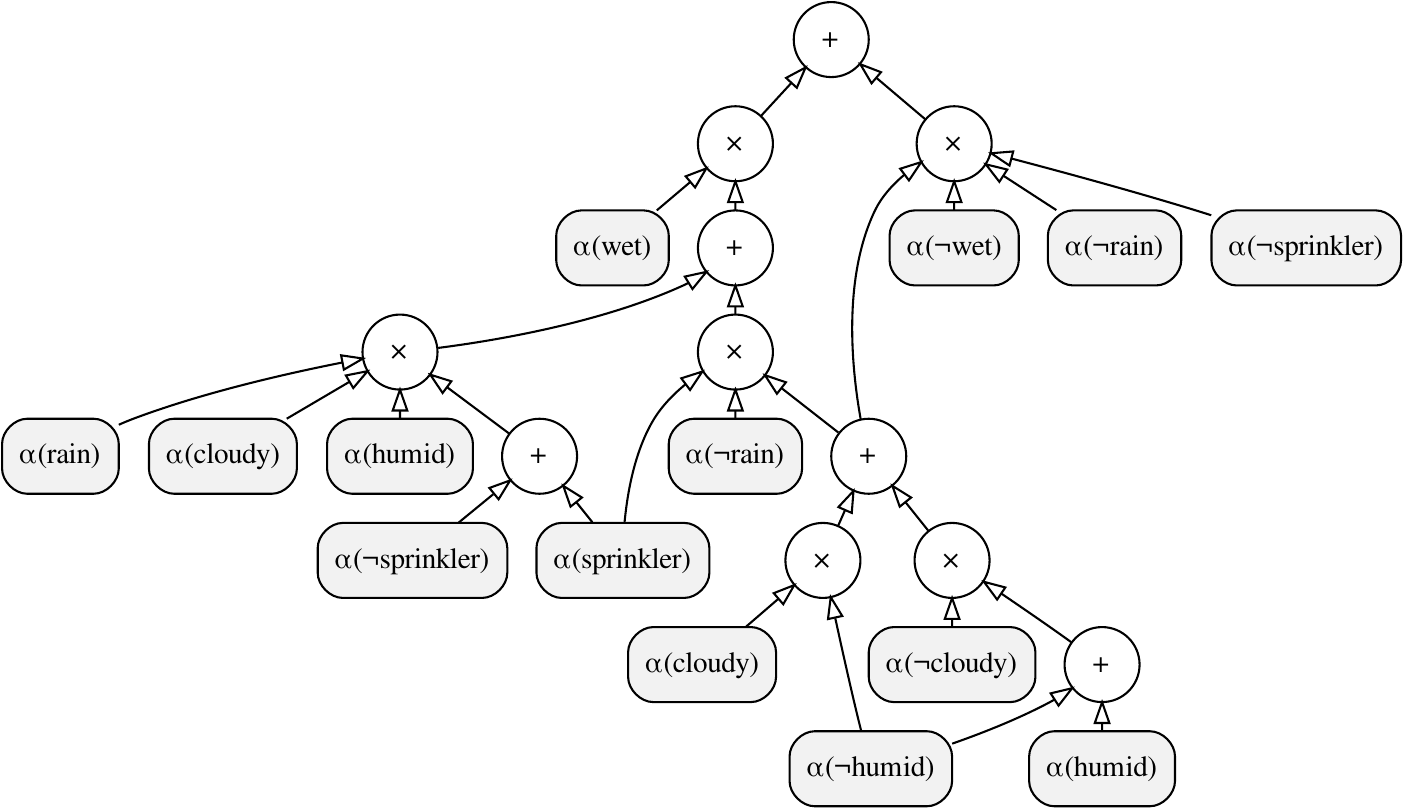}
    \caption{A $WMC(\theory)$ representation of Figure~\ref{fig:sd_ddnnf}.}
    \label{fig:sd_ddnnf_wmc}
\end{figure}

\begin{example}[sd-DNNF]
    Figure~\ref{fig:sd_ddnnf} shows the ProbLog program of Example~\ref{ex:bayesian_network} in sd-DNNF, the theory $\theory$ of which was already illustrated in Example~\ref{ex:wmc}. The corresponding representation of $WMC(\theory)$ is illustrated in Figure~\ref{fig:sd_ddnnf_wmc}.
\end{example}

%
%
Crucially the procedure's soundness relies on the sd-DNNF properties, and on the commutative semiring properties of $(\mathcal{A},\land, \lor, \top, \bot)$.
Consequently, the same compilation procedure can be used more generally to compute an algebraic model count $AMC(\theory, \semiring, \alpha)$. 
When converting into a computational graph, we then use $\oplus$, $\otimes$, $e^\oplus$ and $e^\otimes$ instead of $+$, $\times$, $0$ and $1$.

%
%
Once the computational graph $WMC(\theory)$ is available, the count $WMC(\theory, \alpha)$ can be computed by instantiating each $\alpha(l)$ with their value and propagating all values upwards. The result of the root node is exactly $WMC(\theory, \alpha)$. 
Evidently, the second phase of this counting approach is linear in the size of the computational graph. This procedure is similar for $AMC(\theory, \semiring, \alpha)$.

%
%
The compilation approach is not the only possible approach. Instead of explicitly compiling the sd-DNNF, some approaches intertwine both the compilation and evaluation and avoid keeping the entire compiled structure in memory. While some sd-DNNF compilers have flags for this behavior, e.g. D4~\citep{LagniezM17}, others are more specific to counting-only~\citep{DudekPV20}. 

A major functional advantage of the compiled approach however is that the computational graph can be re-used for several queries, or with a different $\alpha$. In this way the largest computational cost is amortized, and re-evaluation is linear in the graph size. This is especially useful for parameter learning as then only $\alpha$ varies, e.g. learning in DeepProbLog~\citep{manhaeve2021neural}.

\section{Learning}\label{sec:learning}

The generalisations of the fact as discussed in Section~\ref{sec:facts} can also introduce parameters to the logic program. For probabilistic facts, the probability itself can be learned, and would thus be a parameter of the model. Similarly, the distributions in the distributional facts can have learnable parameters, e.g. the mean and standard deviation of a normal distribution. For the neural predicate, the weights of the neural networks are usually considered to be learnable paremeters.

We now consider the parameter learning setting. Given a program with parameters $\mathcal{W}$, a set $\mathcal{Q}$ of tuples $(G,p)$ with $G$ a query and $p$ the target probability, and a loss function $\mathcal{L}$, compute the following Equation: 
\begin{equation}
    \argmin_{\mathcal{W}} \frac{1}{|\mathcal{Q}|}\sum_{(G,p)\in \mathcal{Q}} \mathcal{L}(P_{\mathcal{W}}(G),p).
    \label{eq:loss_function_eq}
\end{equation}

Earlier approaches to parameter learning used an expectation-maximization approach.
Recently, gradient-based optimization has become the dominant strategy for learning. AMC enables us to automatically derive gradients for the parameters in the program through the use of the gradient semiring, which we explain in Section~\ref{sec:gradient_semiring}. After the gradients have been calculated, standard gradient-based optimizers can be used.
When the parameters are contained in differentiable structures (e.g. in a neural network), they are easy to optimize in conjunction with other parameters, as the same gradient-based techniques can be used.

\subsection{Gradient semiring} \label{sec:gradient_semiring}
To derive gradients, we  use  the gradient semiring \citep{KimmigBR11}. The elements of this semiring are tuples 
\[\left(p,\frac{\partial p}{\partial w}\right) \text{,}\]
where $p$ is a probability, and $\frac{\partial p}{\partial w}$ is the  partial derivative of that probability with respect to a parameter $w$.
This is easily extended  to a vector of parameters $\vec{w} = [w_1,\ldots,w_N]^T$, the concatenation of all $N$ parameters in the ground program. The elements of the semiring then become tuples 
\[
\left(p, \nabla p\right) \text{,}
\]
where $p$ is a probability and $\nabla p$ the gradient of $p$ with respect to all parameters in $\vec{w}$.
Addition~$\oplus$, multiplication~$\otimes$ and the neutral elements with respect to these operations are defined as follows:
\begin{align}
(p_1, \nabla p_1) \oplus (p_2, \nabla p_2) &= (p_1+p_1, \nabla p_1 + \nabla p_2),  \\
(p_1, \nabla p_1) \otimes (p_2, \nabla p_2) &= (p_1p_2, p_2 \nabla p_1 +  p_1 \nabla p_2),\\
e^\oplus &= (0,\vec{0}),\\
e^\otimes &= (1,\vec{0}).
\end{align}

Note that the first element of the tuple performs ProbLog's probability computation, whereas the second element computes the gradient of the first element. 

To perform parameter learning in ProbLog, we use the mapping $\alpha$ defined below, where the vector $\mathbf{e}_i$ has a $1$ in the i-th position and $0$ in all others:
\begin{align}
\alpha(f) &=  (p,\vec{0}) &&\text{for}~p\prob f~\text{with fixed}~p, \\
\alpha(f_i) &=  (p_{i}, \mathbf{e}_i) &&\text{for}~t(p_i)\prob f_i~\text{with learnable}~p_i, \\
\alpha(\neg f) &= (1-p, - \nabla p) &&\text{with}~\alpha(f) = (p, \nabla p).
\end{align}

\begin{figure}[ht!]
    \centering
    \includegraphics[width=0.9\linewidth]{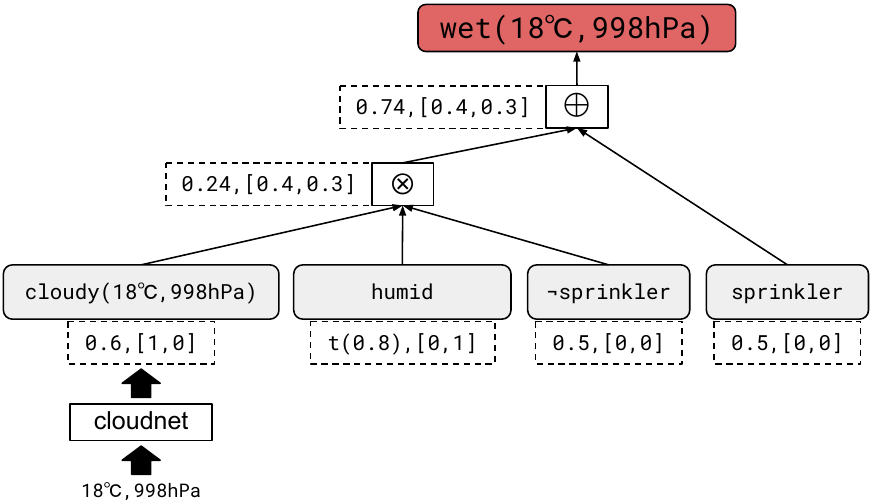}
    \caption{The arithmetic circuit for Example~\ref{ex:learning}. Each node is annotated with an element of the gradient semiring, where the two elements of the gradient represent the partial derivative of the probability with respect to the neural network, and the probabilistic parameter respectively.}
    \label{fig:sprinkler_ac}
\end{figure}

\begin{example}\label{ex:learning}
We demonstrate the joint learning of probabilistic parameters and the neural network's parameters for the program in Example~\ref{ex:neuralfact}. For this example, we learn the parameter of \probloginline{humid} and jointly train the \texttt{cloudnet} network. We use cross-entropy as our loss function $\mathcal{L} = -(p \log(P(G)) + (1-p) \log(1-P(G)))$, with $p$ the target probability and $P(G)$ the probability predicted using the current weight parameters. Since the target probability is $1$, $\mathcal{L}$ is $-\log(P(G))$. 
To update the parameters, we must compute
\[
\frac{\partial\mathcal{L}}{\partial\mathcal{W}}=\frac{\partial\mathcal{L}}{\partial P(G)}\frac{\partial P(G)}{\partial\mathcal{W}} = \frac{-1}{P(G)}\frac{\partial P(G)}{\partial\mathcal{W}} \text{.}
\]
To compute this gradient, we need both $P(G)$ and $\frac{\partial P(G)}{\partial\mathcal{W}}$, which we can calculate using the gradient semiring. The resulting arithmetic circuit after all stages of inference is given in Figure~\ref{fig:sprinkler_ac}. The final gradients are thus $\frac{\partial\mathcal{L}}{\partial\mathtt{humid}} = -0.41$ and $\frac{\partial\mathcal{L}}{\partial\mathtt{cloudy(18^{\circ}C,998~hPa)}} = -0.54$. 
\end{example}

\section{Related Work and Applications} \label{sec:related}
There a many more variants of the logical fact introduced in related work. We discuss a few more in this section. DTProbLog \citep{BroeckTOR10} is a decision-theoretic variant of ProbLog which adds the decision fact, along with the possibility of assigning utilities to atoms. The probabilistic facts, rules, and utilities define an expected utility given a set of values for the decision facts. These values thus define a decision problem that needs to be solved in order to maximize the expected utility.
BetaProbLog~\citep{VerreetDMR22} further generalizes the concept of the probabilistic fact by replacing the single probability (i.e. a point estimate) by a beta distribution which additionally models the epistemic uncertainty over the probability of the fact. 
NeurASP~\citep{yang2020neurasp} takes the idea of the neural fact, and applies it to answer set programming, a different logic programming language, where it is called the neural atom.
In \citet{belle2020semiring}, the authors introduce Semiring Programming, a declarative framework where semirings are used to model and solve a wide variety of tasks in AI.
In Scallop~\citep{huang2021scallop}, semirings are used to define approximate inference in order to alleviate the intractability that methods such as DeepProbLog encounter.
In \citep{zuidberg2021neural}, the authors investigate the setting in which the semiring operations are functions that have to be learned.
Other work involving semirings and the application of them onto functional programming, include \citep{Dolan13}~and~\citep{vandenBerg2023}.

An application domain well suited for algebraic and probabilistic logic programming is in the field of robotics. Here, the probabilistic modelling and reasoning capabilities have been used for object tracking, affordance and object manipulation~\citep{MoldovanORMS11,MoldovanMOSR12,MoldovanAH12,MoldovanMNSR18,NittiLR14,AntanasMNFKSR19,PerssonMRL20,MartiresNPLR20}, for representing and tracking cognitive knowledge about the environment~\citep{Veiga2019AHA,Mekuria2019APM,Yang2023}, and for performing the uncertain decision making itself~\citep{BroeckTOR10,NittiBR15,NittiBLR17,Bueno2016MarkovDP,LatourBDKBN17,derkinderen2020algebraic,venturato2022towards}.

Other application tasks include activity recognition~\citep{SkarlatidisAFP15,McAreaveyBLH17,SztylerCS18,SmithPB21}, consistency-based diagnosis~\citep{HommersomB16}, modeling incomplete and imprecise information~\citep{DohertyS22}, system prognostics~\citep{vlasselaer2012statistical}, ontology matching and querying~\citep{Wang15a,BremenDJ19,BremenDJ20}, probabilistic argumentation~\citep{Hung17,Mantadelis2020ProbabilisticAA,Totis2021CoRR}, solving word-problems~\citep{DriesKDBR17,SusterFTKDRD21}, automating video montages~\citep{aerts2016probabilistic}, epidemiological modelling~\citep{WeitkamperSS21}, game-playing~\citep{Thon2008,ThonLR11}, event processing~\citep{XingVGCKPS19,ApricenoPS21,RoigVilamala2023}, modelling probabilistic routing networks~\citep{van2021probabilistic}, and biology~\citep{Raedt07,Kimmig2012,DeMaeyer2013,DeMaeyer2015,DeMaeyer2016,Alexander2019}.

\section{Conclusion} \label{sec:conclusion}
In this paper, we have shown a unified perspective that describes how logic programming can be extended to a wide variety of settings by generalizing the concept of the fact. We have shown how these extensions are all special cases of the concept of the algebraic fact, where facts are labeled with elements from commutative semirings, and the conjunction and disjunction are replaced with multiplication and addition respectively. We have further shown a recipe for efficient inference and learning for programs that include such algebraic facts.
Finally, we have discussed other works that perform similar extensions, and where these systems have been applied.
Going forward, it would be valuable to look into what benefits the use of semirings has for other languages and programming paradigms. We have already identified links to functional programming, and new, semiring-focused declarative frameworks.

\section*{Acknowledgements}

This work was supported by the Research Foundation-Flanders (FWO) under grant 1SA5520N and S007318N, the KU Leuven Research fund, the Flemish Government under the ``Onderzoeksprogramma Artifici\"ele Intelligentie (AI) Vlaanderen'' programme, 
and the Wallenberg AI, Autonomous Systems and Software Program (WASP) funded by the Knut and Alice Wallenberg Foundation.

\bibliography{main_preprint}

\end{document}